\documentclass[10pt,twocolumn,letterpaper]{article}

\usepackage{cvpr}
\usepackage{times}
\usepackage{epsfig}
\usepackage{graphicx}
\usepackage{tabularx}
\usepackage{multirow}
\usepackage{amsmath}
\usepackage{amssymb}
\usepackage{bm}
\usepackage{bbm}
\usepackage{caption}
\usepackage{algorithm}
\usepackage{algorithmic}
\usepackage{booktabs}
\usepackage{enumitem}

\usepackage{array}
\newcolumntype{?}{!{\vrule width 1pt}}
\newcolumntype{C}[1]{>{\centering\arraybackslash\hspace{0pt}}p{#1}}

\usepackage{color}

\newif\ifdraft
\draftfalse

\newcommand{\comment}[1]{}

\definecolor{olive}{RGB}{50,150,50}

\ifdraft
 \newcommand{\ms}[1]{{\color{olive}{#1}}}
 
 \newcommand{\ar}[1]{{\color{blue}{#1}}}
 \newcommand{\pf}[1]{{\color{red}{#1}}}
 \newcommand{\PF}[1]{{\color{red}{\bf #1}}}
 \newcommand{\MS}[1]{{\color{olive}{\bf #1}}}
  \newcommand{\AR}[1]{{\color{blue}{\bf #1}}}
 \newcommand{\HR}[1]{{\color{magenta}{ #1}}}
\else
 \newcommand{\ms}[1]{#1}
 
 \newcommand{\ar}[1]{#1}
 \newcommand{\pf}[1]{#1}
 \newcommand{\AR}[1]{}
 \newcommand{\PF}[1]{}
 \newcommand{\MS}[1]{}
 \newcommand{\HR}[1]{}
\fi

\newcommand{\ResNet}{\textsc{ResNet}}

\newcommand{\relu}[0]{\texttt{ReLU}}

\newcommand{\df}[0]{\mathop{=}\limits_{df}}

\newcommand{\bTheta}{\mathbf{\Theta}}
\newcommand{\bGamma}{\mathbf{\Gamma}}
\newcommand{\balpha}{\mathbf{\mathcal{A}}}
\newcommand{\bbeta}{\mathbf{\mathcal{B}}}
\newcommand{\bbias}{\mathbf{\mathcal{D}}}
\newcommand{\cN}{\sigma}

\newcommand{\sT}{\mathcal{T}}
\newcommand{\sL}{\mathcal{L}}
\newcommand{\sZ}{\mathcal{Z}}

\newcommand{\bA}{\mathbf{A}}
\newcommand{\bB}{\mathbf{B}}
\newcommand{\bb}{\mathbf{b}}
\newcommand{\bd}{\mathbf{d}}
\newcommand{\bW}{\mathbf{W}}
\newcommand{\bx}{\mathbf{x}}
\newcommand{\by}{\mathbf{y}}

\newcommand{\fig}[1]{Fig.~\ref{fig:#1}}
\newcommand{\sect}[1]{Section~\ref{sec:#1}}

\newcommand{\tbl}[1]{Table~\ref{tbl:#1}}

\newcommand{\eqt}[1]{Eq.~\ref{eq:#1}}

\newcommand{\alg}[1]{Algorithm~\ref{alg:#1}}

\newcommand{\norm}[1]{\left\lVert#1\right\rVert}
\newcolumntype{M}[1]{>{\centering\arraybackslash}m{#1}}
\newcolumntype{R}[1]{>{\raggedleft\arraybackslash}m{#1}}
\newcolumntype{P}[1]{>{\centering\arraybackslash}p{#1}}

\newcommand{\argmin}{\operatornamewithlimits{argmin}}

\usepackage[pagebackref=true,breaklinks=true,letterpaper=true,colorlinks,bookmarks=false]{hyperref}

\cvprfinalcopy 

\ifcvprfinal\fi
\begin{document}

\title{Residual Parameter Transfer for Deep Domain Adaptation}

\author{Artem Rozantsev \qquad Mathieu Salzmann \qquad Pascal Fua \\
  Computer Vision Laboratory, \'{E}cole Polytechnique F\'{e}d\'{e}rale de Lausanne\\
  Lausanne, Switzerland\\
  {\tt\small \{firstname.lastname\}@epfl.ch}
}

\maketitle

\begin{abstract}

The goal of Deep Domain Adaptation is to make it possible to use Deep Nets trained in one domain where there is enough annotated training data in another where there is little or none. Most current approaches have focused on learning feature representations that are invariant to the changes that occur when going from one domain to the other, which means using the same network parameters in both domains. While some recent algorithms explicitly model the changes by adapting the network parameters, they either severely restrict the possible domain changes, or significantly increase the number of model parameters.

By contrast, we introduce a network architecture that includes auxiliary residual networks, which we train to predict the parameters in the domain with little annotated data from those in the other one. This architecture enables us to flexibly preserve the similarities between domains where they exist and model the differences when necessary. We demonstrate that our approach yields higher accuracy than state-of-the-art methods without undue complexity.

\end{abstract}

\comment{ allows using the power of deep neural networks on the datasets with very little or no available annotations by leveraging the informaiton from a different dataset that has lots of them.}
\comment{\pf{Deep Domain Adaptation is a powerful way to adapt \MS{In fact, most methods do not really adapt the parameters (unless you think of pre-training on the source domain only, which is typically the case.} the parameters learned by a deep network }}


\section{Introduction}

Given enough training data, Deep Neural Networks~\cite{Hinton06a,LeCun98b} have proven extremely powerful. However, there are many situations where sufficiently large training databases are difficult or even impossible to obtain.  In such cases, Domain Adaptation~\cite{Jiang08} can be used to leverage annotated data from a {\it source domain} in which it is plentiful to help learn the network parameters in a {\it target} domain in which there is little, or even no, annotated data. 

The simplest \ar{approach to Domain Adaptation} is to use the available annotated data in the target domain to fine-tune a Convolutional Neural Network (CNN) pre-trained  on the source data~\cite{Girshick14,Oquab14}. However this can result in overfitting when too little labeled target data is available and is not applicable at all in the absence of any such labeled target data. 

One way to overcome this problem is to design features that are invariant to the domain shift, that is, the differences between the statistics in the two domains. This is usually done by introducing loss terms that force the statistics of the features extracted from both domains to be similar~\cite{Tzeng14,Long15b,Ganin16,Tzeng15}. While effective when the domain shift is due to lighting or environmental changes, enforcing this kind of statistical invariance may discard information and negatively impact performance.
\begin{figure}[!t]
	\centering
	\includegraphics[width = \linewidth]{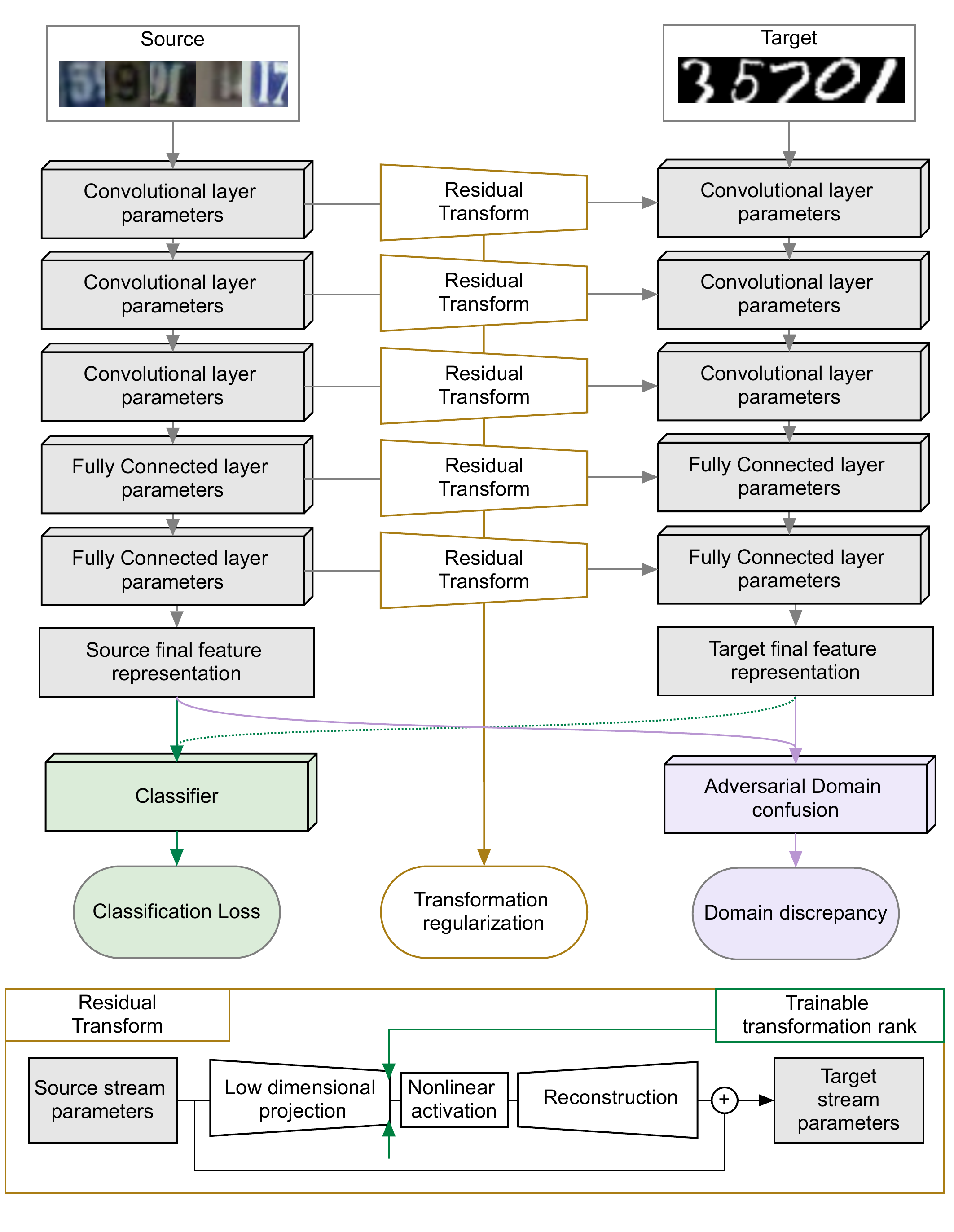}
	\vspace{-9mm}
	\caption{{\bf Our  two-stream architecture.} One stream  operates on
		the source data  and the other on the target  one. Their parameters are
		{\it not} shared. Instead, we introduce a residual transformation network that relates the parameters of the streams with each other.}
	\label{fig:teaser}
\end{figure}
To overcome this, it has been proposed to explicitly model the domain shift~\cite{Bousmalis16, Rozantsev16b}. In particular, the method in~\cite{Bousmalis16} involves learning private and shared encoders for each of the domains, which increases the number of parameters to be learned by a factor of $4$. By contrast, the approach in~\cite{Rozantsev16b} relies on a two-stream architecture with related but non-shared parameters to model the shift. This only require a $2$-fold increase in the number of parameters to be learned but at the cost of restricting corresponding parameters in the two domains to approximately be scaled and shifted versions of each other. Furthermore, it requires selecting the layers that have non-shared parameters using a validation procedure that does not scale up to modern very deep architectures such as those of~\cite{Simonyan15,He16}.

In this paper, we also explicitly model the domain shift between the two domains using a two-stream architecture with non-shared parameters. However, we allow for a much broader range of transformations between the parameters in both streams and automatically determine during training how strongly related corresponding layers should be. As a result, our approach can be used in conjunction with very deep architectures. 

Specifically, we start from a network trained on the source data and 
\ms{fine-tune it while learning additional} auxiliary, residual networks that adapt the layer parameters to make the final target feature distribution as close as possible to the final distribution of the source features. Furthermore, we regularize the capacity of these auxiliary networks \ms{by finding an optimal rank for their parameter matrices, and thus learn the relationship between corresponding layers in the two streams.} Our contribution therefore is twofold:
\begin{itemize}

	\item We model the domain shift by learning meta parameters that transform the \comment{parameters}\ar{weights and biases} of each layer of the network. They are depicted by the horizontal branches in \fig{teaser}.
	
	\item We propose an automated scheme to adapt the complexity of these transformations during learning. \comment{by finding an optimal rank for the weight matrices that control their behavior.}
	
\end{itemize}
This results in a performance increase compared to the approaches of~\cite{Bousmalis16} and~\cite{Rozantsev16b}, along with a reduction in the number of parameters to be learned by a factor $2.5$ and $1.5$ compared to the first and second, respectively. As demonstrated by our experiments, we also outperform the state-of-the-art methods that attempt to learn shift invariant features~\cite{Ganin16,Long17}.


\section{Related Work}
\label{sec:related}

\comment{
Approaches to Domain Adaptation (DA) can be classified in two main categories: the classical ones~\cite{Saenko10,Fernando13, Ghifary14, Chen16c} and the ones based on deep learning algorithms. In this section we will focus on the second class of methods, as they are the most relevant to our approach and show empirical superiority with respect to the first class of methods. 
}

Most approaches to domain adaptation (DA) that operate on deep networks focus on learning features that are invariant to the domain shift~\cite{Tzeng14,Long15b,Ganin16,Tzeng15,Long16b,Long17a}. This is usually achieved by adding to the loss function used for training a term that forces the distributions of the features extracted from the source and target domains to be close to each other. 

In~\cite{Tzeng14}, the additional loss term is the Maximum Mean Discrepancy (MMD) measure~\cite{Gretton07}. This was extended in~\cite{Long15b} by using multiple MMD kernels to better model differences between the two domains. In~\cite{Long17a},
this was further extended by computing the loss function of~\cite{Long15b} at multiple levels, including on the raw classifier output.
The MMD measure~\cite{Gretton07} that underpins these approaches is based on first order statistics. This was later generalized to second-order statistics~\cite{Sun16,Sun16c} and to even higher-order ones~\cite{Koniusz17}.

In~\cite{Ganin16,Tzeng15,Ganin17} a different approach was followed, involving training an additional classifier to predict from which domain a sample comes. \ms{These methods then aim} 
to learn a feature representation that fools this classifier, or, in other words, that carries no information about the domain a sample belongs to. This adversarial approach eliminates the need to manually model the distance measure between the final source and target feature distributions and enables the network to learn it automatically.

While effective, all these methods aim to learn domain invariant features, with a single network shared by the source and target data. By contrast, in~\cite{Tzeng17}, a network pre-trained on the source domain was refined on the target data by minimizing the adversarial loss of~\cite{Tzeng15} between the fixed source features and the trainable target representation.
Furthermore, in~\cite{Bousmalis16}, differences and similarities between the two domains are modeled separately using private and shared encoders that generate feature representations, which are then given to a reconstruction network. The intuition is that, by separating domain similarities and differences, the network preserves some information from the source data and learns the important properties of the target data. While effective, this quadruples the total number of model parameters, thus restricting the applicability of this approach to relatively small architectures. In the same spirit, the approach of~\cite{Rozantsev16b} relies on a two-stream architecture, one devoted to each domain. Some layers do not share their parameters, which are instead encouraged to be scaled and shifted versions of each other. While effective, this severely restricts the potential transformations from one domain to the other. Furthermore, the subsets of layers that are shared or stream-specific are found using a validation procedure, which scales poorly to the very deep architectures that achieve state-of-the-art performance in many applications. 

In this paper, we introduce a two-stream architecture that suffers from none of these limitations.

\begin{figure*}[!ht]
	\centering
	\includegraphics[width=0.97\linewidth]{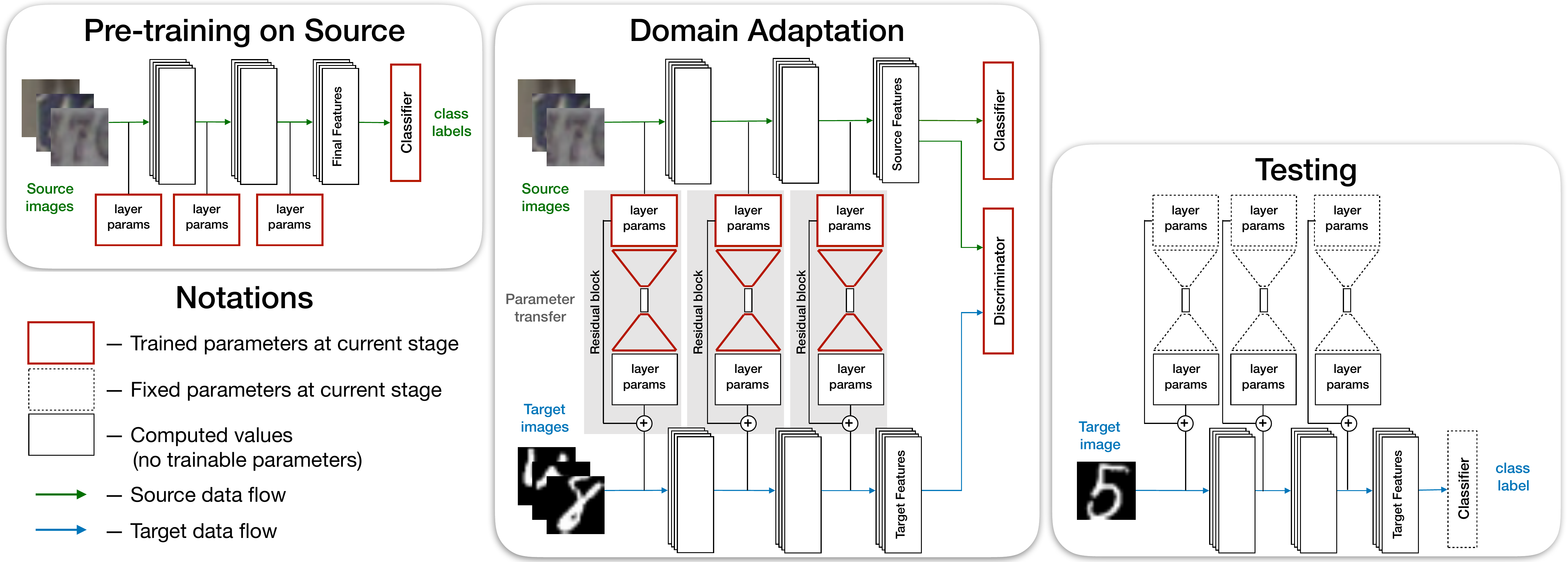}
        \vspace{-2mm}
	\caption{{\bf Approach overview.} We first pre-train the network on the source data. We then jointly learn the source stream parameters and their transformations using adversarial domain adaptation. Finally, at test time, we use the network with transformed parameters to predict the labels of images from the target domain. (Best seen in color)}
	\label{fig:algo}
\end{figure*}

\section{Approach}
\label{sec:approach}

We start from an arbitrary network that has been trained on the source domain, which we refer to as source stream. We then introduce auxiliary networks that transform the source stream parameters to generate a target stream, as depicted by \fig{teaser}. We jointly train the auxiliary networks and refine the original source stream using annotated source data and either a small amount of annotated samples from the target domain or unlabeled target images only. We refer to the former as the {\it supervised} case and to the latter as the {\it unsupervised} one.

\fig{algo} summarizes our  approach. Its Domain Adaptation component appears in the center, and we now describe it in detail. To this end, we first introduce our auxiliary networks and then show that we can control the rank of their weight matrices to limit the number of parameters that need to be learned. In effect, during training, our network automatically learns which layers should be different from each other and which ones can have similar or equal parameters. 

\subsection{Adapting the Parameters of Corresponding Layers}

Let  $\Omega$ be the set of all layers in a single stream of the Deep Network architecture illustrated in \fig{teaser}. For each layer  $i \in\Omega$, let us first consider a vector representation of the source and target stream parameters as $\theta_i^s$ and $\theta_i^t$, respectively.

\comment{$\theta_i^s$ and $\theta_i^t$ be the \ar{parameter} vectors that define the source and the target streams, respectively.  As discussed below, we have experimented with two different ways of writing $\theta_i^t$ as a function of $\theta_i^s$ in terms of matrix multiplications and non linear operations, which can then be used to define auxiliary networks such as the one shown at the bottom of  \fig{teaser}. }

A natural way to transform the source parameters into the target ones is to write
\begin{equation}
\theta_i^{t} = \bB_i\cN(\bA_i^{\intercal} \theta_i^{s} + \bd_i) + \theta_i^{s}, \quad \forall i \in \Omega \;,
\label{eq:deep_weights}
\end{equation}
\noindent
for which the notation is given in \tbl{notations_simple}. Note that $k_i$, the second dimension of the $\bA_i$ and  $\bB_i$  matrices, controls the complexity of the transformation by limiting the rank of the matrices. $k_i=0$ corresponds to the degenerate case where the parameters of the source and target streams are identical, that is, shared between the two streams.


\begin{table}[!h]
	\begin{tabularx}{\linewidth}{r|l|X}
	\toprule
	$\cN(\cdot)$ & $\in \{ \tanh , \relu \}$ & nonlinear activation \\ 
	$\bA_i, \bB_i$ & $\in \mathbb{R}^{M_i \times k_i}$ & transformation matrices\\
	$k_i $ &  $k_i \ge 0$ & transformation rank \\
	$\bd_i $ & $\in \mathbb{R}^{k_i}$ & bias term\\ 
	\midrule
	$M_i$ & \multicolumn{2}{l}{the number of parameters in the $i^{th}$ layer} \\
	$\Omega$ & \multicolumn{2}{l}{the set of all network layers }\\
	\bottomrule
	\end{tabularx}
\caption{Notation for \eqt{deep_weights}.}
\label{tbl:notations_simple}
\end{table}

\comment{
For notational simplicity, we group the individual equations of Eq.~\ref{eq:deep_weights} into a single one
\begin{equation}
\bTheta^{t} = \cN(\balpha \bTheta^{s} + \bbias)\bbeta^{\intercal} + \Theta^{s}
\label{eq:deep_weights_ezy}
\end{equation}
where we omit the index $i$ and \ar{define
\begin{equation}
\small
\balpha \df \{\alpha_i\}, \; \bbeta \df \{\beta_i\}, \; \bbias \df \{d_i\}, \; \bTheta^s \df \{\theta^s_i\}, \; \bTheta^t \df \{\theta^t_i\}. \;
\label{eq:notations_p2}
\end{equation}
}
}

In theory, we could learn all the coefficients of these matrices for all layers, along with their rank, by minimizing a loss function such as the one defined in \sect{fixed}. Unfortunately, the formulation of \eqt{deep_weights} results in a memory intensive implementation because each increase of the transformation rank $k_i$ by $1$ in any layer creates $(2M_i+1)$ additional parameters, which quickly becomes impractically large, especially when dealing with fully-connected layers. 

To address this issue, we propose to rewrite the layer parameters in matrix form. Our strategy for different layer types is as follows:
\begin{itemize}

 \item Fully connected layer. Such a layer performs a transformation of the form $\by = \sigma(\bA \bx +\bb)$, where $\bA$ is a matrix and $\bb$ a vector whose size is the number of rows of $\bA$. In this case, we  simply concatenate  $\bA$ and $\bb$ into a single matrix.
 
 \item Convolutional layer. Such a layer is parametrized by a tensor $\bW \in \mathbb{R}^{N_{out}\times N_{in} \times f_x \times f_y}$, where the convolutional kernel is of size $f_x \times f_y$, and a bias term $\bb \in \mathbb{R}^{N_{out}}$. We therefore represent all these parameters as a single matrix by reshaping the kernel weights as an $N_{out}\times N_{in} f_x f_y$ matrix and again concatenating the bias with it. 
\end{itemize}
Following these operations, the parameters of each layer $i$ in the source and target streams are represented by matrices $\Theta_i^{s}$ and $\Theta_i^{t}$, respectively.  We then propose to write the transformation from the source to the target parameters as
\begin{equation}
\Theta_i^{t} = \bbeta_i^1\cN\hspace{-0.1cm}\left(\left(\balpha_i^1\right)^{\intercal} \Theta_i^{s} \balpha_i^2 + \bbias_i\right)\left(\bbeta_i^2\right)^{\intercal} + \Theta_i^{s} \; ,
\label{eq:deep_weights_v2}
\end{equation}
\noindent
for which the notation is defined in \tbl{params_v2}. This formulation is preferable to the one of  \eqt{deep_weights} because $\balpha_i^1$, $\balpha_i^2$, $\bbeta_i^1$, and $\bbeta_i^2$ are \ms{small compared to $\bA_i$ and $\bB_i$.} \ms{This can be best seen when formally computing the number of additional parameters for every layer $i \in \Omega$. When using \eqt{deep_weights}, this number is $(2N_iC_i + 1)k_i$. In the case of \eqt{deep_weights_v2}, it becomes $2(N_ir_i + C_il_i) + r_il_i$. Provided that $\{l_i,r_i,k_i\}$ are of the same magnitude, and typically much smaller than $\{N_i,C_i\}$, \eqt{deep_weights_v2} results in significantly fewer parameters than \eqt{deep_weights}.}

In practice, to further reduce the number of parameters, we limit the $\balpha$ and $\bbeta$ matrices to being block diagonal so that, for each pair of corresponding layers, the weights are linear combinations of\comment{convolution} weights and the biases of biases. Note that, now, the complexity of the source-target transformation depends on the values $l_i$ and $r_i$. 


\begin{table}[!t]
	\begin{tabularx}{\linewidth}{r|l|X}
		\toprule
		$\Theta^{s}_i$& $\in \mathbb{R}^{C_i \times N_i}$ & source stream parameters \\
		$\Theta^{t}_i$& $\in \mathbb{R}^{C_i \times N_i}$ & target stream parameters \\
		\midrule
		$\balpha_{i}^1, \bbeta_{i}^1$& $ \in \mathbb{R}^{C_i \times l_i}$ & \multirow{3}{*}{transformation parameters} \\
		$\balpha_{i}^2, \bbeta_{i}^2$& $ \in \mathbb{R}^{N_i \times r_i}$ & \\
		$\bbias_i$& $ \in \mathbb{R}^{l_i \times r_i}$ & \\
		\midrule
		$N_i$&$ i \in \Omega$ & number of inputs in $\Theta_i$ \\
		$C_i$&$ i \in \Omega$ & number of outputs in $\Theta_i$ \\
		$l_i$&$ i \in \Omega$ & left transformation rank for $\Theta_i$ \\
		$r_i$&$ i \in \Omega$ & right transformation rank for $\Theta_i$ \\
		$\Omega$& -- & set of network layers \\
		\bottomrule 
	\end{tabularx}
	\caption{Notation for \eqt{deep_weights_v2}.}
	\label{tbl:params_v2}
\end{table}


\subsection{Fixed Transformation Complexity}
\label{sec:fixed}

Let us assume that the $\Theta_i^s$ parameters of the source network have been trained using a standard approach. To achieve our goal of finding the best possible $\Theta_i^t$s, we use \eqt{deep_weights_v2} to express them as functions of the $\Theta_i^s$s, \ms{and define a loss function $\sL(\{\Theta_i^s\},\{\Theta_i^t\})$ that we minimize with respect to} both the source stream parameters $\{\Theta_i^s\}$ and the parameters that define the mapping from the source to the target weights
\begin{equation}
\bGamma = \{\{\balpha_i^1\},\{\balpha_i^2\},\{\bbeta_i^1\},\{\bbeta_i^2\},\{\bbias_i\}\}\;.
\label{eq:transfParams}
\end{equation}

Let us further assume that the potential complexity of the transformation between the source and target domains is known {\it a priori}, that is, the values $l_i$ and $r_i$ are given, an assumption that we will relax in \sect{complexityReduc}. Under these assumptions, we write our loss function as 
\begin{equation}
\sL_{\texttt{fixed}}  = \sL_{\texttt{class}} + \sL_{\texttt{disc}} + \sL_{\texttt{stream}} \; ,
\label{eq:cost_function}
\end{equation}
and describe its three terms below. 

\paragraph{Classification Loss:} $\sL_{\texttt{class}}$. The first term in \eqt{cost_function} is the sum of standard cross-entropy classification losses, computed on the annotated samples from the source and target domains. If there is no annotated data in the target domain, we use the classification loss from the source domain only.

\paragraph{Discrepancy Loss:} $\sL_{\texttt{disc}}$. This term aims to measure how statistically dissimilar the feature vectors computed from the source and target domains are. Minimizing this discrepancy is important because the feature vectors produced by both streams are fed to the {\it same} classifier, as shown in \fig{teaser}.  Ideally, the final representations of the samples from both domains should be statistically indistinguishable from each other. To this end, we take $\sL_{\texttt{disc}}$  to be the adversarial domain confusion loss of~\cite{Ganin16}, which is easy to implement and has shown state-of-the-art performance on a wide range of domain adaptation benchmarks.

Briefly, this procedure relies on an auxiliary classifier $\phi$ that aims to recognize from which domain a sample comes, based on the feature representation learned by the network. $\sL_{\texttt{disc}}$ then favors learning features that fool this classifier. In a typical adversarial fashion, the parameters of the classifier and of the network are learned in an alternating manner. More formally, given the feature representation $\textbf{f}$, the parameters $\theta^{DC}$ of the classifier are found by minimizing the cross-entropy loss
\begin{equation}
\small
\sL_{DC}(y_n) = -\frac{1}{N}\sum_{n = 1}^{N}[ y_n\log(\hat{y}_n) + (1-y_n)\log(1-\hat{y}_n)] \; ,
\label{eq:dc_loss}
\end{equation}
\noindent
where $N$ is the number of source and target samples, $y_n \in [0,1]$ is the domain label, and $\hat{y}_n = \phi(\theta^{DC}, \textbf{f}_n)$. We then take the domain confusion term of our loss function to be 
\begin{equation}
\sL_{\texttt{disc}} = \sL_{DC}(1 - y_n) \; .
\label{eq:dc_rev_loss}
\end{equation}

\paragraph{Stream Loss:}  $\sL_{\texttt{stream}}$. The third term in \eqt{cost_function} serves as a regularizer \ar{to the residual part of the transformation defined in \eqt{deep_weights_v2}}. We write it as 
\begin{equation}
\sL_{\texttt{stream}} = \lambda_s \left(\sL_{\omega} - \sZ \left(\sL_{\omega}\right)\right),
\label{eq:stream}
\end{equation}
\noindent
where
\begin{equation}
\sL_{\omega} = \sum_{i\in \Omega}\norm{\bbeta_i^1\cN\hspace{-0.1cm}\left(\left(\balpha_i^1\right)^{\intercal} \Theta_i^{s} \balpha_i^2 + \bbias_i\right)\left(\bbeta_i^2\right)^{\intercal}}_{Fro}^2 \; .
\label{eq:weight_reg}
\end{equation}
\noindent
$\lambda_s$ controls the influence of this term and $\sZ$ is a barrier function~\cite{Nocedal06}, which we take to be $\log(\cdot)$ in practice. Since $\sL_\texttt{stream}$ is smallest when $\sL_{\omega} = 1$ and goes to infinity when $\sL_{\omega}$ becomes either very small or very large, it serves a dual purpose. First, it prevents the network from learning the trivial transformation $\sL_{\omega} \equiv 0$.
Second, it prevents the source and target weights to become too different from each other and thus regularizes the optimization. As will be shown in  \sect{exp}, we have experimented with different values of $\lambda_s$ and found the results to be insensitive to its exact magnitude. However, setting it to zero quickly leads to divergence and failure to learn the correct parameter transformations. In practice, we therefore set $\lambda_s$ to 1.

\subsection{Automated Complexity Selection}
\label{sec:complexityReduc}

In the previous section, we assumed that the values $l_i$ and $r_i$ defining the shape of the transformation matrices were given. These parameters are task dependent and even though it is possible to manually tune them for every layer of the network, it
is typically suboptimal, and even impractical for truly deep architectures. Therefore, we now introduce additional loss terms that enable us to find the $l_i$s and $r_i$s automatically while optimizing the network parameters. As discussed below, these terms aim to penalize high-rank matrices.
To this end, let 
\begin{equation}
   \sT_i = \left(\balpha_{i}^1\right)^{\intercal} \Theta^{s}_i \balpha_{i}^2 + \bbias_i \;\;\in \mathbb{R}^{l_i \times r_i}\; ,
 \label{eq:tMat}
\end{equation}
which corresponds to the inner part of the transformation in \eqt{deep_weights_v2}. To minimize the complexity of this transformation, we would like to find matrices $\balpha_i^1$ and $\balpha_i^2$ such that the number of \emph{effective} rows and columns in the transformation matrix $\sT_i$ is minimized. By effective, we mean rows and columns whose $L_2$ norm is greater than a small $\epsilon$, and therefore have a real impact on the final transformation. Given this definition, the non-effective rows and columns can be safely removed without negatively affecting the performance. In fact, their removal {\it improves} performance by enabling the optimizer to focus on relevant parameters and ignore the others. In effect, this amounts to reducing the $(l_i, r_i)$ values.

To achieve our goal, we define a regularizer of the form
\begin{equation}
R_{c}(\{\sT_i\}) = 
\sum_{i \in \Omega} \left( \sqrt{N_i}\sum_{c} \norm{(\sT_i)_{\bullet c}}_2 \right),
\label{eq:group_lasso_reg}
\end{equation}
which follows the group Lasso formalism~\cite{Yuan06b,Alvarez16a},
where the groups for the $i^{th}$ layer, represented by $(\sT_i)_{\bullet c}$, correspond to the columns of the transformation matrix $\sT_i$.

In essence, this regularizer encourages zeroing-out entire columns of $\sT_i$, and thus automatically determines $r_i$, provided that we start with a sufficiently large value. We can define a similar regularizer $R_{r}(\{\sT_i\})$ acting on the rows of $\sT_i$, which thus lets us automatically find $l_i$.

We then incorporate these two regularizers in our loss function, which yields the complete loss
\begin{equation}
\sL = \sL_{\texttt{fixed}} + \lambda_r \left( R_{c} + R_{r} \right)\;,
\label{eq:ar_sel_loss}
\end{equation}
where $\lambda_r$ is a weighting coefficient and $\sL_{\texttt{fixed}}$ is defined in \eqt{cost_function}. As will be shown in \sect{exp}, we have experimented with various values of $\lambda_r$ and found our approach to be insensitive to it within a wide range. However, setting $\lambda_r$ too small or too big will result in preservation of the starting transformation ranks or their complete reduction to zero, respectively. In practice we set $\lambda_r$ to $1$. 
 
\subsubsection{Proximal Gradient Descent}
\label{sec:prox_grad}
 
In principle, given the objective function of Eq.~\ref{eq:ar_sel_loss}, we could directly use backpropagation to jointly learn all the transformation parameters. In practice, however, we observed that doing so results in slow convergence and ends up removing very few columns or rows. Therefore, following~\cite{Alvarez16a}, we rely on a proximal gradient descent approach to minimizing our loss function. 

In essence, we use Adam~\cite{Kingma15} for a pre-defined number of iterations to minimize $\sL_{\texttt{fixed}}$ {\it without} the rank minimizing terms, which gives us an estimate of $\hat{\bGamma}$, and thus of the transformation matrices $\hat{\sT}_i$. We then update these matrices using the proximal operator defined as
\begin{equation}
\small
\sT_i^* = \argmin_{\sT_i} \frac{1}{2t}\norm{\sT_i - \hat{\sT_i}}^2_2 + \lambda_r\left(R_{c}(\sT_i) + R_{r}(\sT_i)\right) \; ,
\label{eq:group_lasso_opt}
\end{equation}
where $t$ is the learning rate. In contrast to~\cite{Alvarez16a}, here, we have two regularizers that share parameters of $\sT_i$. To handle this, we solve Eq.~\ref{eq:group_lasso_opt} in two steps, as
\begin{equation}
\begin{aligned}
\bar{\sT_i} & = \argmin_{\sT_i} \frac{1}{4t}\norm{\sT_i - \hat{\sT_i}}^2_2 + \lambda_r R_{c}(\sT_i) \; ,  \\
\sT_i^*     & = \argmin_{\sT_i} \frac{1}{4t}\norm{\sT_i - \bar{\sT_i}}^2_2 + \lambda_r R_{r}(\sT_i) \; .    \\
\end{aligned}
\label{eq:group_lasso_opt_pract}
\end{equation}
\noindent
for each layer $i$ of the network. As shown in~\cite{Yuan06b}, these two subproblems have a closed-form solution.

Given the resulting $\sT = \{\sT_i\}_{i \in \Omega}$, we need to compute the corresponding matrices $\balpha_{i}^1$, $\balpha_{i}^2$ and $\bbias_i$ for every layer, such that Eq.~\ref{eq:tMat} holds. This is an under-constrained problem, since $l_i$ and $r_i$ are typically much smaller than $N_i$ and $C_i$. Therefore, we set $\bbias_i$ to the value obtained after the Adam iterations, and we compute $\balpha_{i}^1$ and $\balpha_{i}^2$ such that they remain close to the Adam estimates and satisfy Eq.~\ref{eq:tMat} in the least-squares sense. We observed empirically that this procedure stabilizes the learning process. More details are provided in the Supplementary Appendix~\ref{sec:LS_sol}. \alg{optimization} gives an overview of our complete optimization procedure.


\begin{algorithm}[!t] 
	\begin{algorithmic}[1]
		\item[] \textbf{Input: }
		\begin{enumerate}[topsep=0cm, leftmargin=0.55cm]
			\setlength\itemsep{0pt}
			\setlength{\parskip}{0pt}
			\item   The two-stream architecture depicted by \fig{teaser}
			\item   Randomly initialized transformation parameters $\bGamma^0$
		\end{enumerate}
		\item[] \textbf{Output: } 
		\begin{enumerate}[topsep=0cm, leftmargin=0.55cm]
			\setlength\itemsep{0pt}
			\setlength{\parskip}{0pt}
			\item   $\{\Theta^s_i\}$ -- the parameters of the source stream
			\item   $\bGamma^{\texttt{res}}$ -- parameters of the auxiliary networks
		\end{enumerate}
		\item[]
		\FOR {$\texttt{epoch} < N_{\texttt{epochs}}$}
		\STATE $\{\Theta^s_i\}, \hat{\bGamma} \leftarrow N$ steps of Adam to minimize $\sL_{\texttt{fixed}}$ 
		\STATE $\{\hat{\sT}_i\} \leftarrow 
		\begin{cases}
		\{\{\hat{\balpha}_i^1\},\{\hat{\balpha}_i^2\},\{\hat{\bbias}_i\} \} \subset \hat{\bGamma} \\
		\{\Theta^s_i\} \\
		\end{cases}$
		 \hfill (\eqt{tMat})
		\STATE $\{\sT_i\} \leftarrow$ group sparse projection of $\{\hat{\sT}_i\}$ \hfill (\eqt{group_lasso_opt})
		\STATE $\{(\balpha_i^1,\balpha_i^2)\} \leftarrow$ LS estimate from $\{\sT_i\}$ 
		\STATE $\bGamma^{\texttt{epoch}} \leftarrow \{\{\balpha_i^1\},\{\balpha_i^2\},\{\hat{\bbeta}_i^1\},\{\hat{\bbeta}_i^2\},\{\hat{\bbias}_i\} \}$
		\ENDFOR 
		\STATE $\bGamma^{\texttt{res}} \leftarrow \bGamma^{N_{\texttt{epochs}}}$
	\end{algorithmic}
	\caption{Optimization Procedure}
	\label{alg:optimization}
\end{algorithm}

\section{Experiments}
\label{sec:exp}

In this section, we first discuss the baseline methods that we used in our experiments.
We then compare our approach to them in three very different contexts, hand-written character recognition, drone detection, and office object recognition\ms{, further demonstrating} that our approach applies to very deep architectures such as \ResNet{}s~\cite{He16}. 


\subsection{Baseline Methods}

As discussed in Section~\ref{sec:related}, deep domain adaptation techniques can be roughly classified into those that attempt to learn features that are invariant to the domain change and those that modify the weights of the network that operates on the target data to take into account the domain change. 

The approach of~\cite{Ganin16} is an excellent representative of the first class. Furthermore, since we incorporate its adversarial domain confusion term $\sL_{\texttt{disc}}$ into our own loss function, it makes sense to use it as a baseline to gauge the increase in performance our complete framework brings about. 

Our own method belongs to the second class of which the works of~\cite{Bousmalis16, Rozantsev16b,Tzeng17} are the most recent representatives. We therefore also use them as baselines. 


\subsection{SVHN to MNIST: Unsupervised Adaptation}

\begin{figure}[!t]
	\centering
	\begin{tabular}{cccccccc}
		\toprule
		\multicolumn{8}{c}{SVHN} \\
					   \includegraphics[width=0.1\linewidth]{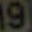} &
		\hspace{-0.3cm}\includegraphics[width=0.1\linewidth]{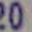} &
		\hspace{-0.3cm}\includegraphics[width=0.1\linewidth]{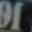} &
		\hspace{-0.3cm}\includegraphics[width=0.1\linewidth]{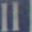} &
		\hspace{-0.3cm}\includegraphics[width=0.1\linewidth]{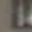} &
		\hspace{-0.3cm}\includegraphics[width=0.1\linewidth]{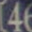} &
		\hspace{-0.3cm}\includegraphics[width=0.1\linewidth]{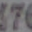} &
		\hspace{-0.3cm}\includegraphics[width=0.1\linewidth]{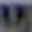} \\
					   \includegraphics[width=0.1\linewidth]{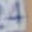} &
		\hspace{-0.3cm}\includegraphics[width=0.1\linewidth]{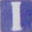} &
		\hspace{-0.3cm}\includegraphics[width=0.1\linewidth]{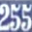} &
		\hspace{-0.3cm}\includegraphics[width=0.1\linewidth]{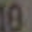} &
		\hspace{-0.3cm}\includegraphics[width=0.1\linewidth]{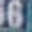} &
		\hspace{-0.3cm}\includegraphics[width=0.1\linewidth]{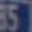} &
		\hspace{-0.3cm}\includegraphics[width=0.1\linewidth]{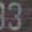} &
		\hspace{-0.3cm}\includegraphics[width=0.1\linewidth]{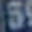} \\
		\midrule
		\multicolumn{8}{c}{MNIST} \\
					   \includegraphics[width=0.1\linewidth]{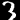} &
		\hspace{-0.3cm}\includegraphics[width=0.1\linewidth]{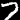} &
		\hspace{-0.3cm}\includegraphics[width=0.1\linewidth]{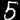} &
		\hspace{-0.3cm}\includegraphics[width=0.1\linewidth]{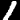} &
		\hspace{-0.3cm}\includegraphics[width=0.1\linewidth]{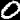} &
		\hspace{-0.3cm}\includegraphics[width=0.1\linewidth]{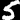} &
		\hspace{-0.3cm}\includegraphics[width=0.1\linewidth]{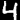} &
		\hspace{-0.3cm}\includegraphics[width=0.1\linewidth]{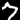} \\
					   \includegraphics[width=0.1\linewidth]{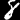} &
		\hspace{-0.3cm}\includegraphics[width=0.1\linewidth]{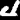} &
		\hspace{-0.3cm}\includegraphics[width=0.1\linewidth]{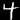} &
		\hspace{-0.3cm}\includegraphics[width=0.1\linewidth]{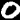} &
		\hspace{-0.3cm}\includegraphics[width=0.1\linewidth]{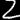} &
		\hspace{-0.3cm}\includegraphics[width=0.1\linewidth]{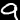} &
		\hspace{-0.3cm}\includegraphics[width=0.1\linewidth]{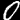} &
		\hspace{-0.3cm}\includegraphics[width=0.1\linewidth]{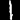} \\
		\bottomrule
	\end{tabular}
	\caption{Images from the SVHN and MNIST domains.}
	\label{fig:svnh2mnist}
\end{figure}

In this section, we analyze our method's unsupervised behavior on the popular SVHN $\rightarrow$ MNIST domain adaptation benchmark for character recognition. As depicted by \fig{svnh2mnist}, SVHN contains images of printed digits while MNIST features hand-written ones.  Following standard practice~\cite{Ganin16,Tzeng17},  we take SVHN to be the source domain and MNIST the target one. 


\subsubsection{Evaluation}
\label{sec:svhn_mnist_eval}

To show that our approach  is not tied to a specific network architecture, we tested two different ones, \textsc{SvhNet}~\cite{Srivastava14} and \textsc{LeNeT}~\cite{LeCun98}, which are the architectures also used  by our baselines~\cite{Ganin16,Tzeng17}. Not only do these architectures have different numbers of convolutional filters and neurons in the fully connected layers, they also work with different image sizes,  $32 \times 32$ for \textsc{SvhNet} and  $28 \times 28$ for  \textsc{LeNeT}.  

In both cases, to test the unsupervised behavior of our algorithm, we used the whole \pf{annotated training set} of SVHN to train the network in the source domain. We then used all the training images of MNIST {\it without}  annotations to perform domain adaptation in an unsupervised manner.\comment{, that is, without the $\sL_{\texttt{class}}$ class term of \eqt{cost_function}.} \tbl{UDA_SVNH} summarizes the results in terms of mean accuracy value and its variance over $5$ runs of the algorithm. From one run to the next, the only difference is the order in which the training samples are considered.
 Our method clearly outperforms the others independently of the architecture we tested it on. 

We also report the results of our approach {\it without} the complexity reduction of Section~\ref{sec:complexityReduc}, that is, by mininizing the loss $\sL_{\texttt{fixed}}$ of \eqt{cost_function} instead of the full loss function $\sL$ of \eqt{ar_sel_loss}. Note that reducing the complexity helps improve performance\comment{, presumably} by reducing the number of parameters that must be learned. In \tbl{svhn_ranks}, we provide
the transformation ranks for each feature-extracting layer of the \textsc{LeNeT} architecture before and after complexity reduction. In this case, the first layer retains its high rank while the others are sharply reduced. This suggests a need to strongly adapt the parameters of the first layer to the new domain, whereas those of the other layers can remain more strongly related to the source stream.

\begin{table}[!t]
	\centering
	\begin{tabularx}{\linewidth}{c|XM{2.5cm}}
		\toprule
		\multicolumn{2}{c}{\phantom{abc}} & \small{SVHN $\rightarrow$ MNIST} \\
		\cmidrule{3-3}
		\multicolumn{1}{l}{\scriptsize{model}} & & \scriptsize{Accuracy: Mean [Std]} \\
		\midrule
		\multirow{5}{*}{\rotatebox{90}{\textsc{SvhNet}}} & 
		Trained on Source data						  							& 54.9 \\
		& DC~\cite{Tzeng14}														& 68.1 [0.03] \\
		& DANN~\cite{Ganin16}													& 73.9 [0.79] \\
		& Ours*:~\scriptsize{$\sL_{\texttt{fixed}}$, no layers shared} 			& 77.8 [0.09] \\
		& Ours																	& \textbf{78.7 [0.12]} \\
		\midrule
		\multirow{5}{*}{\rotatebox{90}{\textsc{LeNeT}}} & 
		Trained on Source data                                                  & 60.1 [1.10] \\
		& DANN~\cite{Ganin16}													& 80.7 [1.58] \\
		& ADDA~\cite{Tzeng17}													& 76.0 [0.18] \\
		& Two-stream~\cite{Rozantsev16b}									    & 82.8 [0.20] \\
		& Ours																    & \textbf{84.7 [0.17]} \\
		\midrule
		custom
		& 
		Domain Separation~\cite{Bousmalis16} 									& 82.78 \\
		\bottomrule 
	\end{tabularx}
	\caption{Comparison to the baseline DA techniques on the SVHN to MNIST benchmark. The accuracy numbers for the baseline methods are taken from the respective papers.}
	\label{tbl:UDA_SVNH}
\end{table}
%

\begin{table}[t!]
	\centering
	\begin{tabular}{cc}
		\hspace{-0.3cm} \raisebox{-2.4cm}{\includegraphics[width=0.33\linewidth]{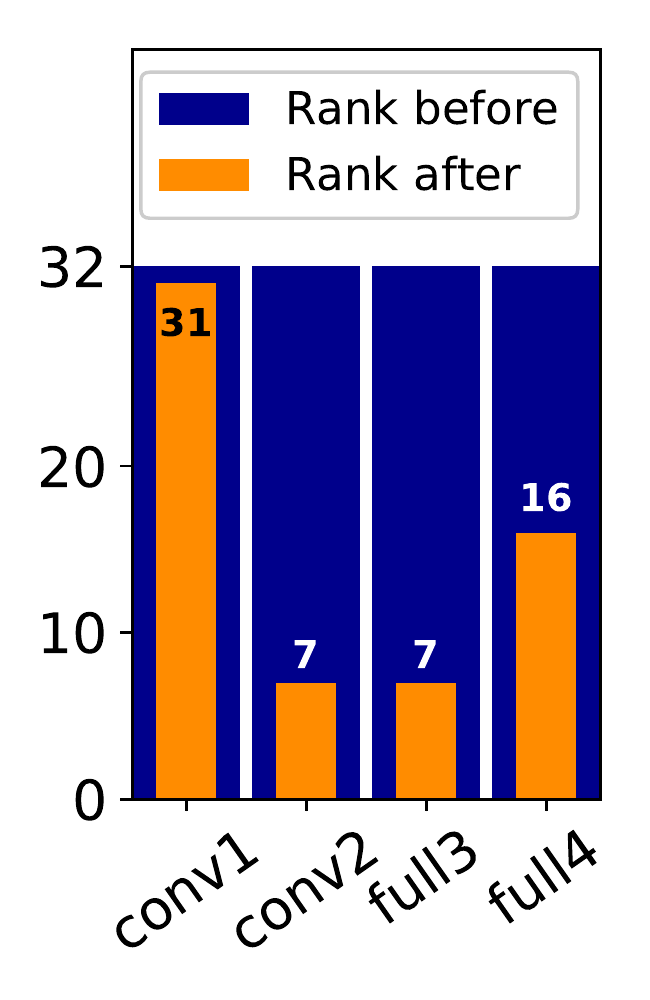}} & \hspace{-0.2cm}
		\begin{tabularx}{0.6\linewidth}{XM{1.5cm}M{1.5cm}}
			\toprule
			& \multicolumn{2}{c}{\small{Transformation ranks: [l, r]}}\\
			\cmidrule{2-3}
			& \scriptsize{before} & \scriptsize{after} \\
			\midrule
			conv1 & [32, 32] & [31, 31] \\
			conv2 & [32, 32] & [9, 7] \\
			full3 & [32, 32] & [7, 7] \\
			full4 & [32, 32] & [16, 16]\\
			\bottomrule
		\end{tabularx} \\
	\end{tabular}
	\vspace{-2mm}
	\caption{Automated complexity selection. [\textsc{Left}] Reduction of the transformation ranks in each \textsc{LeNeT} layer. The layers are shown on the $x$-axis and the corresponding ranks before and after optimization on the $y$-axis. [\textsc{Right}] The same information expressed in terms of the $l_i$ and $r_i$ parameters before and after complexity reduction.}
	\label{tbl:svhn_ranks}
\end{table}
	
\subsubsection{Hyperparameters}

In Section~\ref{sec:approach}, we introduced two hyper-parameters that control the relative influence of the different terms in the loss function of Eq.~\ref{eq:ar_sel_loss}. They are $\lambda_s$, the weight that determines  the influence of the regularization term in Eq.~\ref{eq:stream}, and $\lambda_r$, the weight of Eq.~\ref{eq:ar_sel_loss}, which controls how much the optimizer tries to reduce the complexity of the final network. 

\begin{figure}[!t]
	\centering
	\begin{tabular}{c}
	\includegraphics[width=0.95\linewidth]{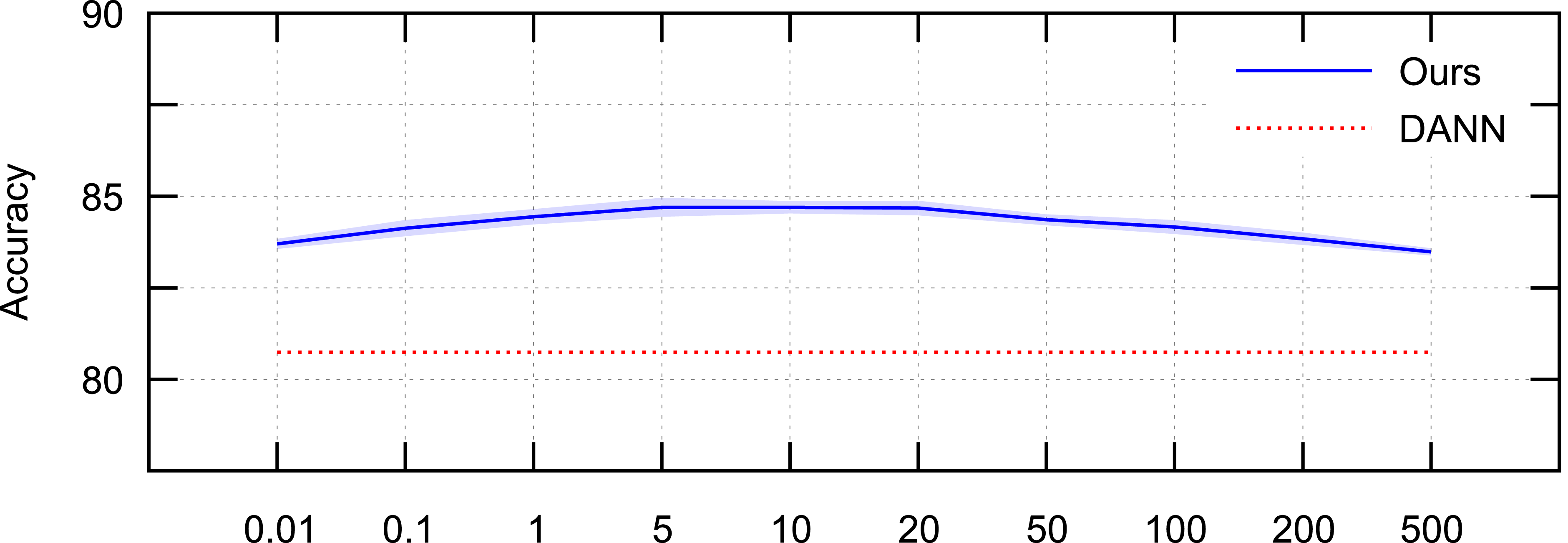} \\
	\scriptsize{Stream Loss weight: $\lambda_s$} \\
	\includegraphics[width=0.95\linewidth]{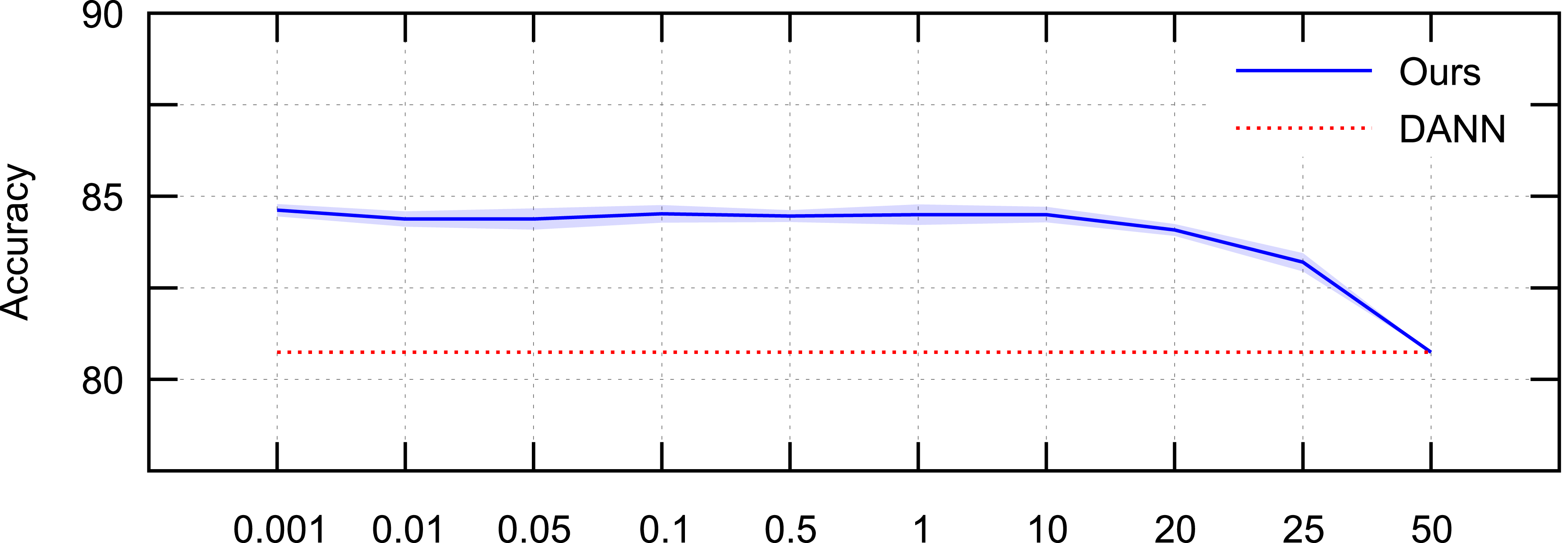} \\
	\scriptsize{Group Sparsity weight: $\lambda_r$} \\
	\end{tabular}
	\vspace{-2mm}
	\caption{Accuracy as a function of the values of hyperparameters $\lambda_s$ and $\lambda_r$ on the SVHN to MNIST benchmark. It is shown in blue and changes little over a wide range. In practice we take  $\lambda_s=\lambda_r=1$. We also plot in red the performance of DANN~\cite{Ganin16} for comparison purposes.}
	\label{fig:hyp_lambda}
\end{figure}

In \fig{hyp_lambda}, we plot the accuracy as a function of $\lambda_s$ and $\lambda_r$. It is largely unaffected over a large range of values, meaning that the precise setting of these two hyper-parameters is not critical. Only when $\lambda_r$ becomes very large do we observe a significant degradation because the optimizer then has a tendency to reduce all transformation ranks to $0$, which means that the source and target stream parameters are then completely shared.
In all other experiments reported in this paper, we set  $\lambda_r$ and $\lambda_s$ \pf{to 1.}

 \comment{

\paragraph{Stream Loss weight:} $\lambda$. Through our experiments we have empirically found that our method is robust to the exact values of these parameter. \fig{hyp_lambda} depicts the mean and standard deviation of the accuracy of our method across $5$ runs for different values of $\lambda$. As we can see the performance of the method remain high in a relatively large interval of values with a slight decrease on the extremes. We further provide the performance of DANN~\cite{Ganin16} for reference.

\paragraph{Group Sparsity weight:} $\lambda_r$. Similarly to the previous case we have found that our method is robust to the exact value for this weighting parameter. As you can see from \fig{hyp_lambda} the accuracy of our method is high for a large range of values. The performance degrades, only if this parameter is set too big, as in this case the regularizer is too strong, which leads to reduction of the transformation ranks to $0$ for all the network layers.

}



\subsection{Drone Detection: Supervised Adaptation}

We now evaluate our approach on the \emph{UAV-200} dataset of~\cite{Rozantsev16b}. It comprises $200$ labeled real UAV images and approximately $33\text{k}$ synthetic ones, which are used as positive examples at training time. It also includes about $190\text{k}$ real images without UAVs, which serve as negative samples. To better reflect a detection scenario, at test time, the quality of the models is evaluated in terms of Average Precision (AP)~\cite{Su15c} on a set of $3\text{k}$ real positive UAV images and $135\text{k}$ negative examples. The training and testing images are of course kept completely separate. 

We treat the synthetic data as the source domain and the real images as the target one. Our goal is therefore to leverage what can be learned from the synthetic images to instantiate the best possible network for real images even though we have very few to train it. In other words, we tackle a need in tasks where synthesizing images is becoming increasingly easy but acquiring real ones in sufficient quantity remains difficult. 


\begin{figure}[!t]
	\centering
	\begin{tabular}{cccccccc}
		\toprule
		\multicolumn{8}{c}{Synthetic} \\
					   \includegraphics[width=0.1\linewidth]{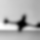} &
		\hspace{-0.3cm}\includegraphics[width=0.1\linewidth]{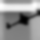} &
		\hspace{-0.3cm}\includegraphics[width=0.1\linewidth]{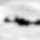} &
		\hspace{-0.3cm}\includegraphics[width=0.1\linewidth]{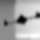} &
		\hspace{-0.3cm}\includegraphics[width=0.1\linewidth]{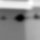} &
		\hspace{-0.3cm}\includegraphics[width=0.1\linewidth]{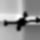} &
		\hspace{-0.3cm}\includegraphics[width=0.1\linewidth]{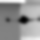} &
		\hspace{-0.3cm}\includegraphics[width=0.1\linewidth]{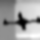} \\
					   \includegraphics[width=0.1\linewidth]{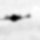} &
		\hspace{-0.3cm}\includegraphics[width=0.1\linewidth]{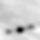} &
		\hspace{-0.3cm}\includegraphics[width=0.1\linewidth]{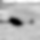} &
		\hspace{-0.3cm}\includegraphics[width=0.1\linewidth]{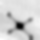} &
		\hspace{-0.3cm}\includegraphics[width=0.1\linewidth]{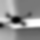} &
		\hspace{-0.3cm}\includegraphics[width=0.1\linewidth]{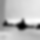} &
		\hspace{-0.3cm}\includegraphics[width=0.1\linewidth]{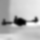} &
		\hspace{-0.3cm}\includegraphics[width=0.1\linewidth]{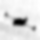} \\
		\midrule
		\multicolumn{8}{c}{Real} \\
					   \includegraphics[width=0.1\linewidth]{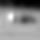} &
		\hspace{-0.3cm}\includegraphics[width=0.1\linewidth]{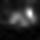} &
		\hspace{-0.3cm}\includegraphics[width=0.1\linewidth]{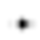} &
		\hspace{-0.3cm}\includegraphics[width=0.1\linewidth]{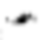} &
		\hspace{-0.3cm}\includegraphics[width=0.1\linewidth]{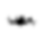} &
		\hspace{-0.3cm}\includegraphics[width=0.1\linewidth]{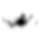} &
		\hspace{-0.3cm}\includegraphics[width=0.1\linewidth]{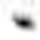} &
		\hspace{-0.3cm}\includegraphics[width=0.1\linewidth]{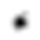} \\
					   \includegraphics[width=0.1\linewidth]{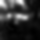} &
		\hspace{-0.3cm}\includegraphics[width=0.1\linewidth]{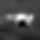} &
		\hspace{-0.3cm}\includegraphics[width=0.1\linewidth]{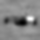} &
		\hspace{-0.3cm}\includegraphics[width=0.1\linewidth]{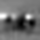} &
		\hspace{-0.3cm}\includegraphics[width=0.1\linewidth]{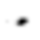} &
		\hspace{-0.3cm}\includegraphics[width=0.1\linewidth]{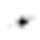} &
		\hspace{-0.3cm}\includegraphics[width=0.1\linewidth]{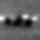} &
		\hspace{-0.3cm}\includegraphics[width=0.1\linewidth]{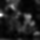} \\
		\bottomrule
	\end{tabular}
	\caption{Synthetic and real UAV images.}
	\label{fig:uav_synth2real}
\end{figure}

We compare our method against several baselines. In \tbl{UDA_UAV}(right), we report the results in terms of mean and standard deviation of the Average Prevision metric across $5$ runs for each method. Ours clearly outperforms the others. It is followed by the two-stream architecture of~\cite{Rozantsev16b} that requires  approximately $1.5$ times as many parameters at training time to perform domain adaptation. \tbl{UDA_UAV}(left) depicts the reduction in transformation ranks that we achieve by automatically learning the complexity of our residual auxiliary networks.

\comment{Why? Was this comment based on fixing the source stream?} 
\comment{$\leftarrow$ no - this is because in~\cite{Rozantsev16b} we double the size of the network at training time, while in our case we have the parameters of the source stream and the parameters of the residual networks. Based on my calculations the ratio turned out to be $1.6$.}

\begin{table}[!t]
	\centering
	\vspace{-0.227cm}
	\begin{tabular}{cc}
	\hspace{-0.4cm} \raisebox{-2.4cm}{\includegraphics[width=0.35\linewidth]{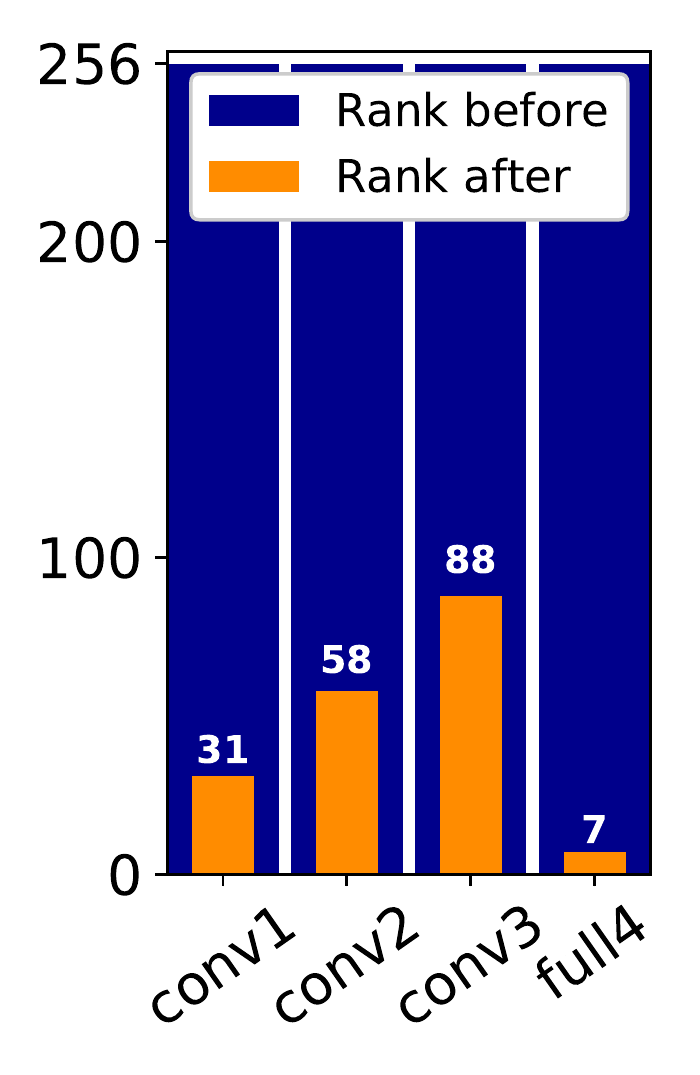}} & 
	\hspace{-0.55cm}
		\begin{tabularx}{0.70\linewidth}{XM{1.7cm}}
			\toprule
			\multicolumn{2}{r}{\small{Synth $\rightarrow$ Real}} \\
			\cmidrule{2-2}
			& \scriptsize{AP: Mean [Std]} \\
			\midrule
			Trained on Source data 						  				& .377 \\
			DANN~\cite{Ganin16}											& .715 [.004] \\
			ADDA~\cite{Tzeng17}											& .731 [.005] \\
			Two-stream~\cite{Rozantsev16b}								& .732 [.003] \\
			Ours														& \textbf{.743 [.006]} \\
			\bottomrule
		\end{tabularx} \\
	\end{tabular}
	\vspace{-2mm}
	\caption{UAV detection. [\textsc{Left}] Reduction of the transformation ranks as in Table~\ref{tbl:svhn_ranks}. [\textsc{Right}] Comparison to baseline domain adaptation techniques.}
	\label{tbl:UDA_UAV}
\end{table}


\subsection{Adaptation with Very Deep Networks}

\begin{figure}[!t]
	\centering
	\begin{tabular}{cccccccc}
		\toprule
		\multicolumn{8}{c}{Amazon (A)} \\
					   \includegraphics[width=0.1\linewidth]{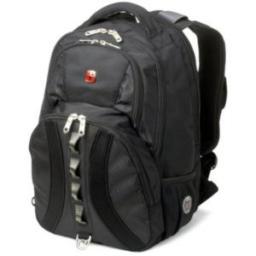} &
		\hspace{-0.3cm}\includegraphics[width=0.1\linewidth]{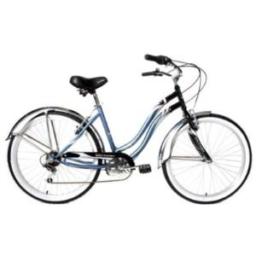} &
		\hspace{-0.3cm}\includegraphics[width=0.1\linewidth]{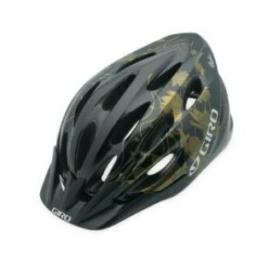} &
		\hspace{-0.3cm}\includegraphics[width=0.1\linewidth]{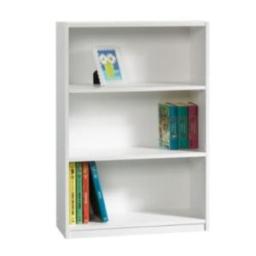} &
		\hspace{-0.3cm}\includegraphics[width=0.1\linewidth]{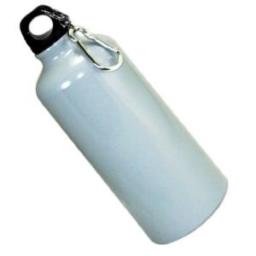} &
		\hspace{-0.3cm}\includegraphics[width=0.1\linewidth]{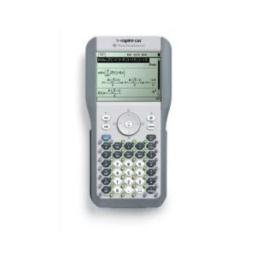} & 
		\hspace{-0.3cm}\includegraphics[width=0.1\linewidth]{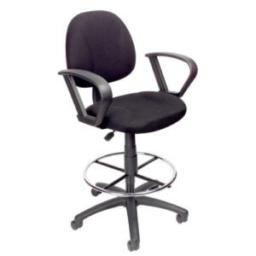} &
		\hspace{-0.3cm}\includegraphics[width=0.1\linewidth]{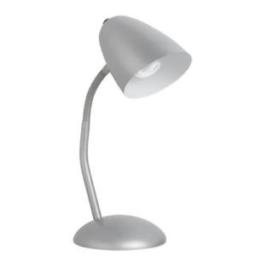} \\
		\midrule
		\multicolumn{8}{c}{DSLR (D)} \\
					   \includegraphics[width=0.1\linewidth]{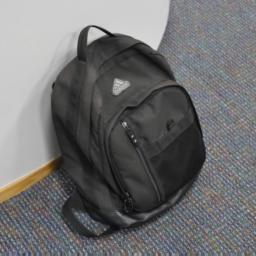} &
		\hspace{-0.3cm}\includegraphics[width=0.1\linewidth]{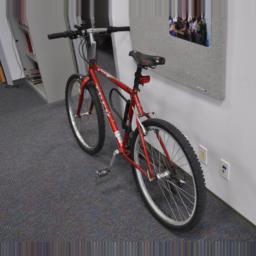} &
		\hspace{-0.3cm}\includegraphics[width=0.1\linewidth]{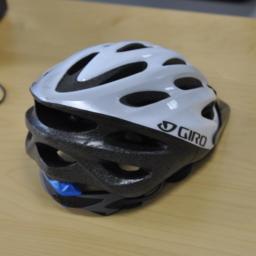} &
		\hspace{-0.3cm}\includegraphics[width=0.1\linewidth]{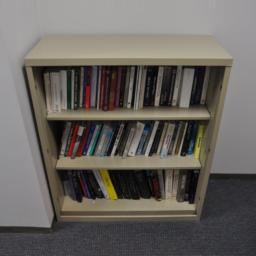} &
		\hspace{-0.3cm}\includegraphics[width=0.1\linewidth]{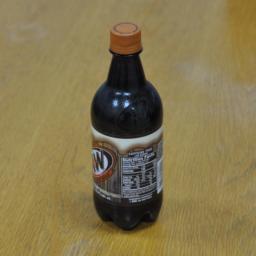} &
		\hspace{-0.3cm}\includegraphics[width=0.1\linewidth]{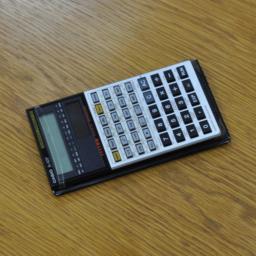} & 
		\hspace{-0.3cm}\includegraphics[width=0.1\linewidth]{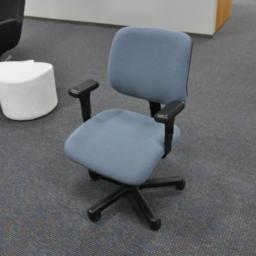} &
		\hspace{-0.3cm}\includegraphics[width=0.1\linewidth]{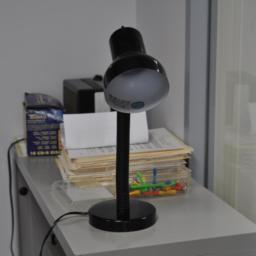} \\
		\midrule
		\multicolumn{8}{c}{Webcam (W)} \\
					   \includegraphics[width=0.1\linewidth]{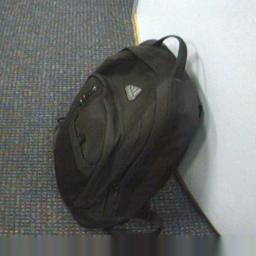} &
		\hspace{-0.3cm}\includegraphics[width=0.1\linewidth]{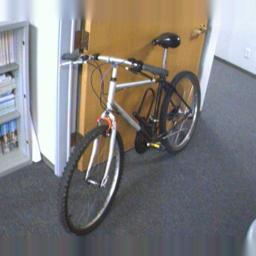} &
		\hspace{-0.3cm}\includegraphics[width=0.1\linewidth]{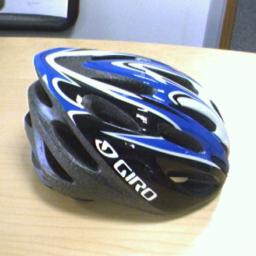} &
		\hspace{-0.3cm}\includegraphics[width=0.1\linewidth]{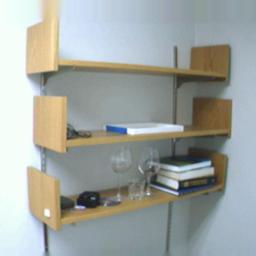} &
		\hspace{-0.3cm}\includegraphics[width=0.1\linewidth]{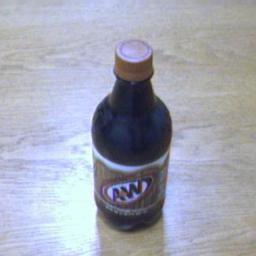} &
		\hspace{-0.3cm}\includegraphics[width=0.1\linewidth]{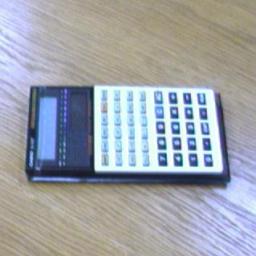} &
		\hspace{-0.3cm}\includegraphics[width=0.1\linewidth]{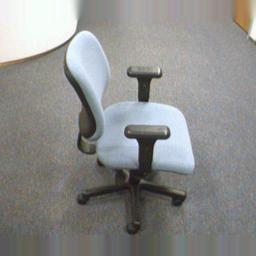} &
		\hspace{-0.3cm}\includegraphics[width=0.1\linewidth]{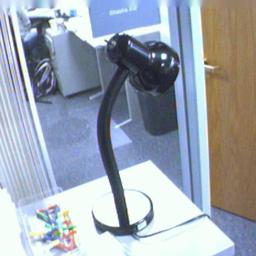} \\
		\bottomrule
	\end{tabular}
	\caption{Sample images from the Office dataset.}
	\label{fig:office}
\end{figure}

To demonstrate that our method can work with very deep architectures, we apply it to the \ResNet-50~\cite{He16} model and analyze its performance on the Office benchmark~\cite{Saenko10} for unsupervised domain adaptation. \fig{office} depicts this dataset. As in~\cite{Long17}, we regularize the final feature representation by adding a `bottleneck' layer to the original \ResNet{} architecture right before the classification layer.  The domain classifier is then connected to the output of this `bottleneck' layer. Since the DANN~\cite{Ganin16} baseline does not use a \ResNet{}, we reimplemented a version of it that does. It relies on the domain confusion network used in~\cite{Ganin16} for the \textsc{AlexNet} model~\cite{Krizhevsky12}.

\comment{I though very few (a single) paper had evaluated with ResNets on Office. How can this be standard practice? Was the same applied to VGG?}
\comment{$\leftarrow$ modified; I have not seen people using VGG for this task, but the works of~\cite{Tzeng14,Ganin16,Long17a} also add a ``bottleneck'' layer after the AlexNet architecture.}

Since the \ResNet{} is very deep, the method of~\cite{Rozantsev16b} that needs to validate all shared/non-shared combinations of layers becomes impractical. Furthermore, the method of~\cite{Bousmalis16} relies on a custom architecture,  which requires increasing the number of parameters by at least a factor of $4$ at training time, thus making it impossible to integrate with \ResNet{} and train on a conventional GPU.
\comment{Is this the true reason, or is it possible but it would lead to an explosion of the number of parameters?} 
\comment{$\leftarrow$ I have modified the sentence a bit.}
Finally, we were unable to make ADDA~\cite{Tzeng17} converge in this case, presumably because when using the \ResNet{} architecture, fine-tuning the target stream with only the domain confusion loss becomes too unconstrained. 

\begin{figure}[!t]
	\centering
	\includegraphics[width=\linewidth]{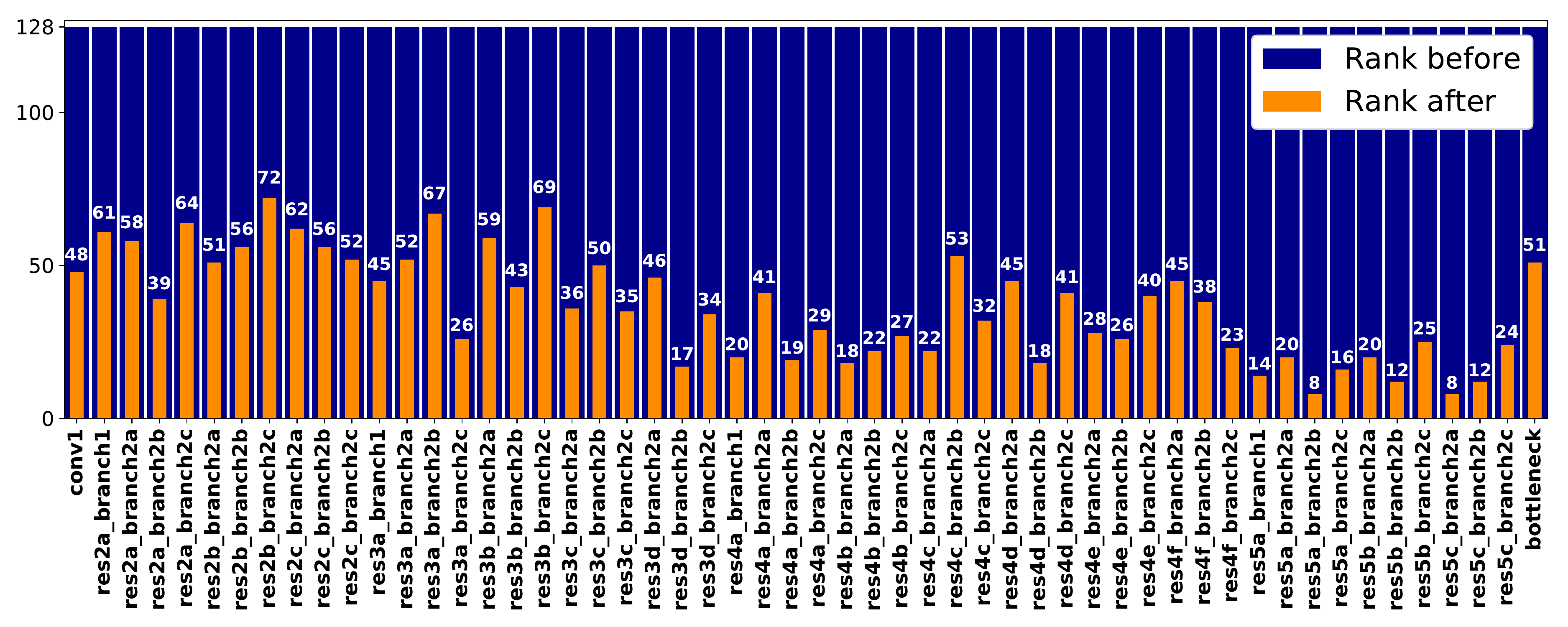}
	\vspace{-5mm}
	\caption{Transformation ranks reduction on the Office dataset for the A$\rightarrow$W domain adaptation task. The ranks significantly shrink for all layers of the \ResNet-50 model. }
	\label{fig:office_ranks}
\end{figure}

In short, our method successfully handles a very difficult domain adaptation task, which creates significant difficulties for all the baselines except 
for DANN~\cite{Ganin16}, which can also deliver results. Nevertheless, as can be seen in \tbl{UDA_OFFICE}, our approach consistently does better.
As before, \fig{office_ranks} illustrates the reduction in complexity that we obtain by automatically adapting the ranks of the residual parameter transformation networks.



\comment{As depicted by  by modeling the difference between the domains we are able to adapt the parameters of the network to the specific properties of the target domain. This allows to improve the accuracy of the classifier with respect to the architecture that extract domain invariant features. \fig{office_ranks} further illustrates that the transformation rank was significantly reduced during the optimization, which shows that our method can be successfully applied to very deep architectures without a significant increase in the number of parameters.
}

\begin{table}[!t]
	\centering
	\begin{tabularx}{\linewidth}{M{0.05cm}M{0.05cm}Xcc}
		\toprule
		\multicolumn{3}{l}{\scriptsize{Domain pair}} & DANN~\cite{Ganin16} & Ours \\
		\midrule
		A &$\rightarrow$& D & 79.1 & \textbf{82.7 [0.3]} \\
		D &$\rightarrow$& A & 63.6 & \textbf{64.7 [0.2]} \\
		A &$\rightarrow$& W & 78.9 & \textbf{81.5 [0.7]} \\
		W &$\rightarrow$& A & 62.8 & \textbf{63.6 [0.2]} \\
		D &$\rightarrow$& W & 97.5 & \textbf{98.0 [0.1]}\\
		W &$\rightarrow$& D & 99.2 & \textbf{99.4 [0.1]}\\
		\bottomrule
	\end{tabularx}
	\vspace{-3mm}
	\caption{Evaluation on the Office dataset using the {\it fully-transductive} evaluation protocol of~\cite{Saenko10}.}
	\label{tbl:UDA_OFFICE}
\end{table}




\section{Conclusion}

We have shown that allowing deep architectures to adapt to the specific properties of the source and target domains improves the accuracy of the final model. To this end, we have introduced a set of auxiliary residual networks that transform the source stream parameters to generate the target stream ones. This, in conjunction with an automatic determination of the complexity of these transformations, has allowed us to outperform the state of the art on several standard benchmark datasets. Furthermore, we have demonstrated that this approach was directly applicable to any network architecture, including the modern very deep ones. In the future, we plan to investigate if adapting the number of layers and of neurons in each layer can further benefit adaptation under more severe domain shifts.

\comment{
We therefore introduced a set of auxiliary residual networks that learn the transformation of the parameters of the source stream to become the parameters of the target stream. This allows to us outperform both methods that learn domain invariant features and state-of-the-art techniques that model the difference between the domains.

In order to remove the need in manually tunning the architectures of the auxiliary networks, we further introduced an approach that automatically learns the complexity of these residual transformations of parameters. In essence, this allows us to model the domain difference with a much smaller number of parameters, with respect to the existing techniques and therefore makes it possible to apply our method to modern very deep architectures.
}

{\small
\bibliographystyle{IEEEtran}
\bibliography{string,vision,learning}
}

\newpage
\section*{Supplementary Appendix}
\appendix


\section{Least-Squares Solution}
\label{sec:LS_sol}

In \sect{prox_grad} we show how to analytically compute $\{\sT_i\}$, that is, the approximation of the solution to \eqt{group_lasso_opt}. Here we discuss in more detail, how to estimate the transformation matrices $\{\balpha^1_i\},\{\balpha^2_i\}$ from the resulting $\{\sT_i\}$. Recall that, for each layer $i \in \Omega$, the inner part of the residual parameter transformation $\sT_i$ is defined as
\begin{equation}
\sT_i = \left(\balpha_{i}^1\right)^{\intercal} \Theta^{s}_i \balpha_{i}^2 + \bbias_i \; , \quad  \sT_i \in \mathbb{R}^{l_i \times r_i} \; .
\label{eq:tMat}
\end{equation} 
\noindent
As discussed in \sect{prox_grad}, we can estimate $\balpha^1_i \in \mathbb{R}^{C_i \times l_i}$ and $\balpha^2_i \in \mathbb{R}^{N_i \times r_i}$ by fixing $\bbias_i$, which in turn allows us to rewrite \eqt{tMat} as 
\begin{equation}
\left(\balpha^1_i\right)^{\intercal}\bTheta^s_i\balpha^2_i = \sT_i - \bbias_i\; ,
\label{eq:LS}
\end{equation}
\noindent
and solve it in the least-squares sense. \eqt{LS}, however, is under-constrained, which leaves us with a wide range of pairs $\{\{\balpha^1_i\}, \{\balpha^2_i\}\}$ that satisfy it. Therefore, in order to make the learning process stable, we suggest finding the optimal $\{\balpha^1_i\}$ and $\{\balpha^2_i\}$ that are the closest to the Adam~\cite{Kingma15} estimates $\{\hat{\balpha}^1_i\},\{\hat{\balpha}^2_i\}$, as discussed in \sect{prox_grad}. To do so, for every layer $i \in \Omega$, we first find the least-squares solution to
\begin{equation}
\small
\balpha^1_i = \argmin_{\tilde{\balpha}^1_i} \norm{\tilde{\balpha}^1_i - \hat{\balpha}^1_i}^2_2 + \norm{\left(\tilde{\balpha}^1_i\right)^{\intercal}\bTheta^s_i\hat{\balpha}^2_i - \sT_i + \bbias_i}^2_2\;,
\label{eq:LS_alpha1}
\end{equation}
\noindent
and then substitute the resulting $\balpha^1_i$ into 
\begin{equation}
\small
\balpha^2_i = \argmin_{\tilde{\balpha}^2_i} \norm{\tilde{\balpha}^2_i - \hat{\balpha}^2_i}^2_2 + \norm{\left(\balpha^1_i\right)^{\intercal}\bTheta^s_i\tilde{\balpha}^2_i - \sT_i + \bbias_i}^2_2\;,
\label{eq:LS_alpha2}
\end{equation}
\noindent
which we then solve in the least-squares sense. As both of these problems are no longer under-constrained, this procedure results in a solution $\{\balpha^1_i\},\{\balpha^2_i\}$ that will both be close to the Adam estimates $\{\{\hat{\balpha}^1_i\},\{\hat{\balpha}^2_i\}\}$ and approximately satisfy \eqt{tMat}. We can then remove the rows of matrix $\balpha^1_i$ and columns of $\balpha^2_i$ that correspond to the rows and columns of $\sT_i$ with an $L_2$ norm less than a small $\epsilon$, as they will make no contribution on the final transformation.

\section{Additional Experiments}

To show that our method does not depend on the specific form of the domain discrepancy loss term $\sL_{\texttt{disc}}$, we have replaced the domain classifier from DANN~\cite{Ganin16} with the one from RMAN~\cite{Long17} that was recently introduced and showed state-of-the-art performance on many Domain Adaptation tasks. In short, this approach builds upon the methods of~\cite{Ganin16} and~\cite{Long17a} by combining the outputs from multiple layers of the feature extracting architecture using random multilinear fusion. 

More formally, in RMAN~\cite{Long17}, the outputs $\{\textbf{f}_j\}$ of a predefined set of layers $\Lambda$ are projected into $D$-dimensional vectors $\{\textbf{p}_j\}$ via a set of random projection matrices $\{\textbf{R}_j\}$. \tbl{notations} describes the notation in more detail. The resulting feature representation $\textbf{f}$ is then formed as
\begin{equation}
\textbf{f} = \frac{1}{\sqrt{D}} \left(\odot_j^{\left|\Lambda\right|} \textbf{p}_j\right) \; ,\;\; j \in \Lambda
\label{eq:product}
\end{equation}
\noindent
where $\odot$ is the element-wise (Hadamard) product. Finally, $\textbf{f}$ is passed to the domain classifier $\phi$, which tries to predict from which domain the sample comes, in the same way as done in~\cite{Ganin16}. We then construct $\sL_{\texttt{disc}}$ in the same manner as in \sect{fixed}. In short, the major difference between DANN~\cite{Ganin16} and RMAN~\cite{Long17} is the input to the domain classifier, which allows for an easy integration of this method with our approach.

\begin{table}[!t]
	\centering
	\begin{tabularx}{\linewidth}{l|l|X}
		\toprule
		$\textbf{f}_j$ & $\in \mathbb{R}^{N_j}$ & layer outputs \\
		$\textbf{p}_j$ & $\in \mathbb{R}^D$ & layer output projections \\
		$\textbf{R}_j$ & $\in \mathbb{R}^{N_j \times D}$ & random projection matrices \\
		$\textbf{f}$   & $\in \mathbb{R}^D$ & resulting feature representation \\
		\midrule
		$N_j$     & \multicolumn{2}{l}{number of output neurons of layer $j \in \Lambda$} \\
		$\Lambda$ & \multicolumn{2}{l}{predefined set of layers} \\
		$D$       & \multicolumn{2}{X}{predefined projection dimensionality} \\
		\bottomrule
	\end{tabularx}
	\caption{Notation}
	\label{tbl:notations}
\end{table}
\begin{figure}[t!]
	\centering
	\includegraphics[width=0.8\linewidth]{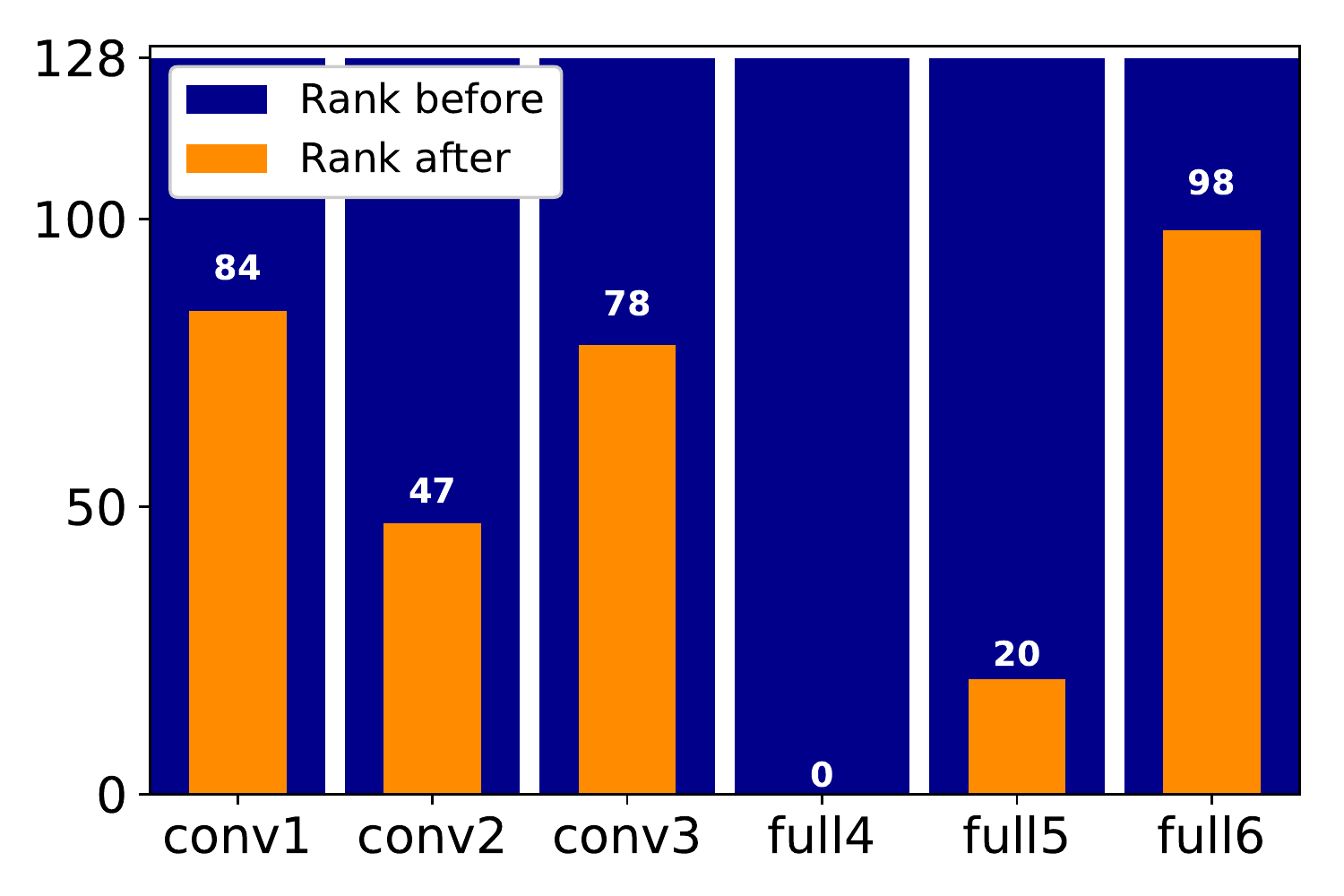}
	\caption{Automated complexity selection. Reduction of the transformation ranks in each \textsc{SvhNet} layer. The layers are shown on the $x$-axis and the corresponding ranks before and after optimization on the $y$-axis.}
	\label{fig:rman_svhn_ranks}
\end{figure}
\begin{table}[t!]
	\centering
	\begin{tabularx}{\linewidth}{Xc}
		\toprule
		& SVHN $\rightarrow$ MNIST \\
		\cmidrule{2-2}
		& \scriptsize{Accuracy: mean [std]} \\
		\midrule
		RMAN~\cite{Long17} 		& 81.0 [0.77] \\
		Ours			   		& \textbf{84.6 [1.26]} \\
		\bottomrule
	\end{tabularx}
	\caption{Comparison of our method that uses the RMAN-based domain discrepancy term with the original RMAN algorithm on the SVHN$\rightarrow$MNIST domain adaptation task.}
	\label{tbl:SVHN_sup_res}
\end{table}

As the code for RMAN is not currently available, we reimplemented it and report the results on the SVHN $\rightarrow$ MNIST domain adaptation task. To this end, we used the \textsc{SvhNet}~\cite{Srivastava14} architecture that was discussed in more detail in \sect{svhn_mnist_eval}. To fuse the output of the final fully-connected layer and the raw classifier output into the $128$-dimensional representation $\textbf{f}$, we used projection matrices $\{\textbf{R}_j\}$, the elements of which were sampled from the a Gaussian distribution $\mathcal{N}(0,1)$. The domain classifier that operates on $\textbf{f}$ has exactly the same architecture as the one used in DANN~\cite{Ganin16} for the SVHN$\rightarrow$MNIST domain adaptation task.

In \tbl{SVHN_sup_res} we compare the results of our approach that uses the RMAN-based domain discrepancy term with the original RMAN method, which can be seen as a version of ours with all the parameters in the corresponding layers shared between the source and target streams. Note that our approach, which allows to transform the source weights to target ones, significantly outperforms RMAN. \fig{rman_svhn_ranks} illustrates the reduction in the complexity of the parameter transfer for every layer of the \textsc{SvhNet} architecture. As in our experiments in \sect{exp}, the ranks of the transformation matrices are significantly reduced during the optimization process.

\end{document}